\pdfoutput=1
\documentclass[11pt]{article}
\usepackage{emnlp2021}
\usepackage{times}
\usepackage{latexsym}
\usepackage[T1]{fontenc}
\usepackage[utf8]{inputenc}

\usepackage{graphicx}
\usepackage{amsmath}
\usepackage{xcolor}
\usepackage{breqn}

\usepackage{booktabs}
\usepackage{multirow}
\usepackage{enumitem}

\usepackage{stfloats}
\usepackage[a-1b]{pdfx}

\usepackage{microtype}

\DeclareMathOperator{\tl}{tl}
\DeclareMathOperator{\tr}{tr}
\DeclareMathOperator{\ro}{ro}
\DeclareMathOperator{\softmax}{softmax}
\DeclareMathOperator{\predict}{predict}
\DeclareMathOperator{\hi}{\text{hi}}
\DeclareMathOperator{\en}{\text{en}}
\DeclareMathOperator{\ml}{\text{ml}}

\title{Cross-Lingual Text Classification of Transliterated \\Hindi and Malayalam}

\author{Jitin Krishnan \quad Antonios Anastasopoulos \quad Hemant Purohit \quad Huzefa Rangwala \\ George Mason University \\ Fairfax, VA, USA \\ 
\texttt{\{jkrishn2,antonis,hpurohit,rangwala\}@gmu.edu}
}

\date{}

\begin{document}
\maketitle
\begin{abstract}

Transliteration 
is very common on social media, but transliterated text is not adequately handled by modern neural models for various NLP tasks. In this work, we combine data augmentation approaches with a Teacher-Student training scheme to address this issue in a cross-lingual transfer setting for \emph{fine-tuning} state-of-the-art pre-trained multilingual language models such as mBERT and XLM-R. 
We evaluate our method on transliterated Hindi and Malayalam, also introducing new datasets for benchmarking on real-world scenarios: one on sentiment classification in transliterated Malayalam, and another on crisis tweet classification in transliterated Hindi and Malayalam (related to the 2013 North India and 2018 Kerala floods). 
Our method yielded an average improvement of $+5.6\%$ on mBERT and $+4.7\%$ on XLM-R in F1 scores over their strong baselines.\footnote{Datasets and implementation available at \url{https://github.com/jitinkrishnan/Transliteration-Hindi-Malayalam}} 

\end{abstract}

\section{Introduction}
\label{intro}

\begin{figure*} [t]
    \centering
    \includegraphics[scale=0.34]{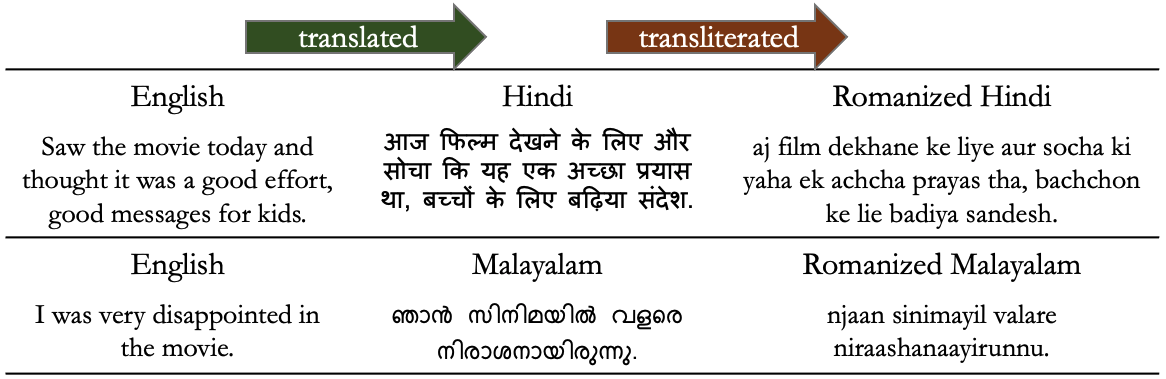}
    \caption{Examples of English sentences and corresponding translations and transliterations. } \label{fig:translit}
\vspace{-1em}
\end{figure*}

A significant number of native Hindi or Malayalam speakers use Latin script instead of Devanagari or other Brahmic scripts for a wide range of social media interactions such as posting tweets, updating Facebook status, commenting on YouTube videos, and writing reviews for restaurants/movies. 
This behavior occurs because keyboard optimizations focus primarily on English \cite{bi2012multilingual} and languages such as Hindi or Malayalam can be very time consuming to type on small devices. If users prefer the original script, the current solution is to back-transliterate the romanized text to the original language \cite{manglishKeyyboard}. Examples of this transliteration process for Hindi and Malayalam are shown in Figure~\ref{fig:translit}. In the first row, English and Hindi are translations of each other, while Latin-transliterated (or \textit{romanized}) 
Hindi is phonetically identical to the Devanagari version but written using Latin characters. 

\begin{figure} 
    \centering
    \includegraphics[scale=0.245]{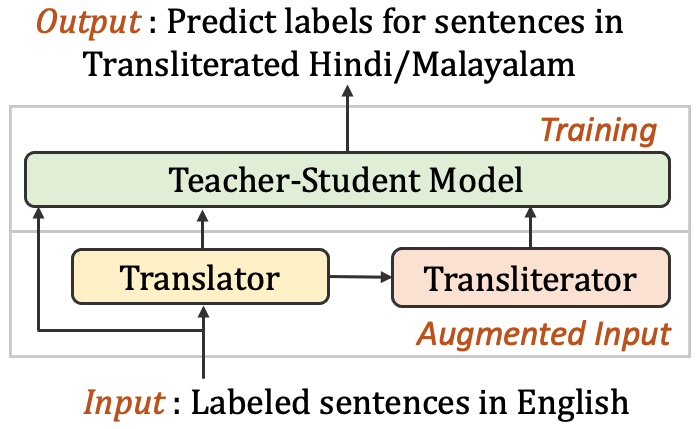}
    \caption{Overview of our method to enhance the multilinguality of transformer models such as mBERT or XLM-R to include Latin-transliterations (romanizations) for two Indic languages: Hindi and Malayalam.} \label{fig:intro}
\vspace{-1em}
\end{figure}

In this context, we explore 
state-of-the-art language models such as multilingual BERT \cite[mBERT]{devlin2018bert} and XLM-R \cite{conneau2019unsupervised} to improve their multilingual generalizability through inclusion of romanized Hindi and Malayalam, as shown in Figure~\ref{fig:intro}. Previous work \cite{pires2019multilingual} has shown that existing transformer-based representations may exhibit systematic deficiencies for certain language pairs.  
This deficiency also appears in transliterated sentences, as shown in the left panels of Figure~\ref{fig:unsup} (in Section \S\ref{results}), where the t-SNE plot of 3-way parallel datasets consisting of sentences in source (English), target (Malayalam/Hindi), and their romanizations are clearly separated in their own clusters, even though they match semantically. We also provide a t-SNE plot of mBERT/XLM-R representations showing this deficiency across various transformer layers in Appendix~\ref{app:A}. 
Our work aims 
to provide 
a general solution towards alleviating this issue, by designing a generic and extensible architecture that can be used for aligning cross-lingual sentence representations. 

Our problem setting is related to language alignment works where static \cite{smith2017offline,conneau2017word} or contextual \cite{aldarmaki2019context,schuster2019cross,cao2020multilingual} word representations from different languages are aligned to a shared vector space. Such methods primarily use a parallel word corpus and either design an alignment loss or explicitly perform rotations (transformations) on the representations. Our work deviates from these methods in three aspects: (a) we focus on sentence-level representation (in this case as produced by the \texttt{[CLS]} classification BERT token; (b) we create a \textit{synthetic 3-way} parallel corpus out of the source data using machine translations and transliterations, and (c) by using a Teacher-Student training scheme, the final representations of the 3 language variants (source, target, and romanized target) are aligned to the same source language space as produced by the original pre-trained model. 

Our Teacher-Student model is inspired from knowledge distillation methods \cite{hinton2015distilling} that are intended to transfer knowledge from a complex model (Teacher) to a simpler model (Student). This approach has been utilized for various tasks, e.g. for reducing the dimension of word embeddings \cite{shin2019pupil}, distilling BERT for text generation \cite{chen2019distilling}, self-knowledge distillation \cite{hahn2019self}, or contrastive learning \cite{chen2020simple,fang2020cert}. We take a similar approach where the Teacher model acts as an anchor by freezing all its layers, while the Student model is fine-tuned based on our optimization procedure to align sentences in English with their target translations and transliterations. Such an approach is necessary, as opposed to other methods like pre-training or self-supervised learning which are outside the scope of this work, because large unlabeled transliterated datasets are not available and collecting them is non-trivial. Thus, we focus on \emph{fine-tuning} and aligning representations of already pre-trained models. 

The practical implications of our work are demonstrated by applying it on naturally-occurring transliterated  datasets of tweets posted during the North India and Kerala flood crises. A model that can immediately handle transliterated tweets by producing embeddings in the same space as that of English tweets can immensely benefit information systems for emergency services, by 
utilizing the vast amount of crisis response models trained in English~\cite{lewis2011crisis,united2020global,nguyen2017robust,krishnan2020unsupervised}. 

\paragraph{Contributions:} \textbf{a)} We propose a novel Teacher-Student method to address the alignment problem for contextual representations of transliterated text produced by multilingual language models (and show its efficacy on Hindi and Malayalam). \textbf{b)} We release two newly labeled datasets; a binary \textit{sentiment} dataset of Malayalam movie reviews and a binary \textit{relevancy} dataset of tweets posted during North India and Kerala flood crises.

\section{Methodology}

We first describe the problem of cross-lingual transfer for transliterations, followed by our Teacher-Student model. 

\subsection{Problem Definition}

Given a source ($S$) dataset in language $\sigma$, e.g. in English (en), the goal is to train a classifier such that it can be used to predict examples from a target ($T$) dataset that consists of transliterations of the target language ($\tau$). To tackle the lack of training data in the transliterated target, as well as the lack of representation alignment between the source sentences $X^\sigma$ and the transliterated target space $X^\tau$, we propose data augmentation. 
Specifically, we first create translations $X^{\tr(\sigma,\tau)}$ of the source language sentences in the target language. Then, we also create transliterations of those translated sentences into the source language's script: $X^{\tl(\tr(\sigma,\tau),\sigma)}$.\footnote{$\tr(a,b)$ represents translation from language $a$ to $b$, and $\tl(a,b)$ represents transliteration from language $a$ to $b$.} All of these are matched with the correct labels $y^\sigma$ from the original dataset. 
Briefly outlined:
\begin{itemize}[noitemsep,leftmargin=*]
    \item[] \textbf{Input:}  $\mathcal{S} = X^{\sigma}$, $y^{\sigma}$
    \item[] \textbf{Augmented Input:}\\   $\mathcal{S'} = X^{\sigma}$, $X^{\tr(\sigma,\tau)}$, $X^{\tl(\tr(\sigma,\tau), \sigma)}$, $y^{\sigma}$
    \item[] \textbf{Goal:} $\mathcal{T} = y^{\tau} \gets \predict(X^{\tau})$.
\end{itemize}


For example, $X^{\tl(\tr(en,hi), en)}$ represents the data that are the result of first translating from English to Hindi, then romanizing the result. The augmentation process can be performed using any existing machine translation and transliteration tool. An example of this process is shown in Figure~\ref{fig:translit}. In the first row, second column is the result of $\tr(\text{en,hi})$ and the third column is the result of $\tl(\tr(\text{en,hi}),\text{en})$. 

\subsection{Teacher-Student Model \& Joint Training}

\begin{figure} 
    \centering
    \includegraphics[scale=0.28]{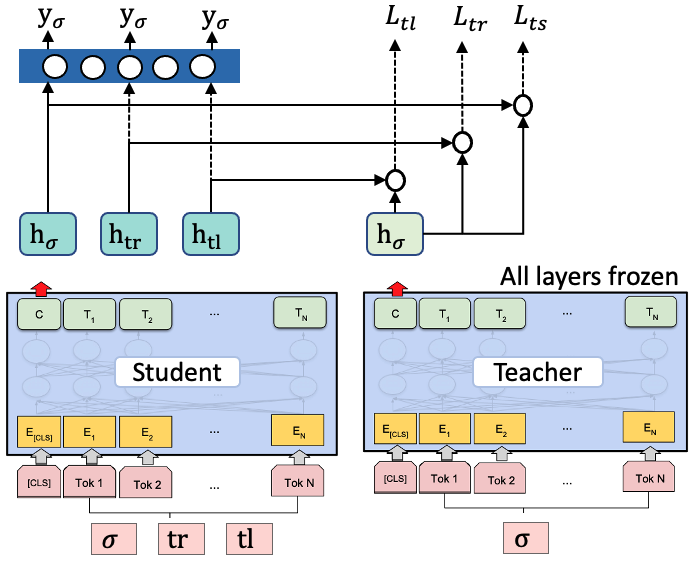}
    \caption{Teacher-Student Model \& Joint Training.} \label{fig:model}
\end{figure}

\begin{figure*} [t]
    \centering
    \includegraphics[scale=0.33]{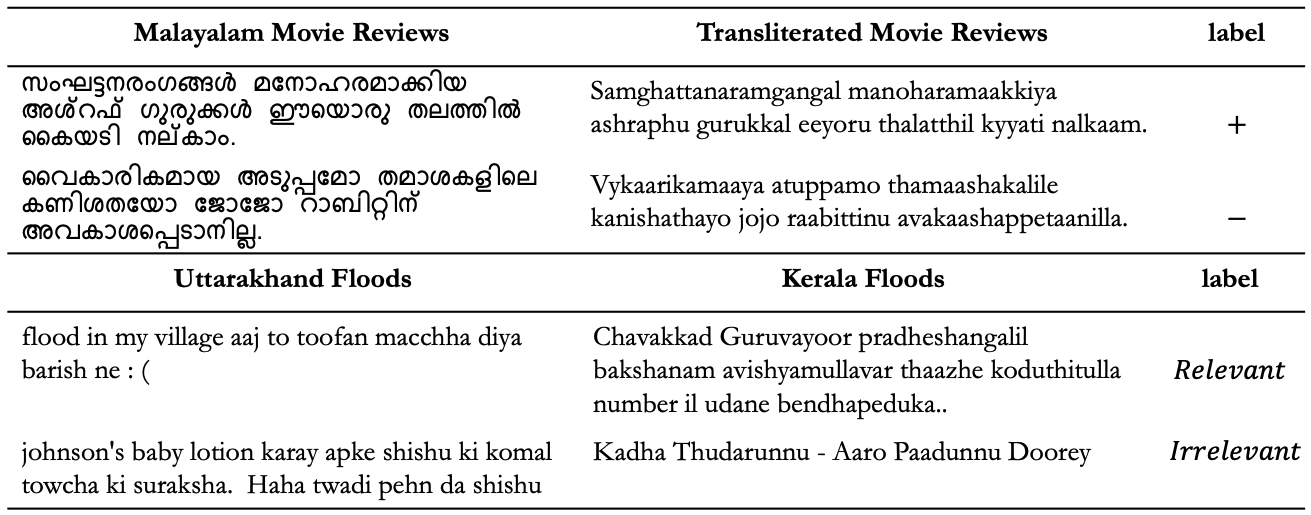}
    \caption{Data samples from new datasets. Malayalam and Transliterated-Malayalam movie sentiment data samples (Top). North India and Kerala floods data samples (Bottom).} \label{fig:ml_stat}
\end{figure*}

The base component of our proposed model is straightforward: it obtains sentence representations from a pre-trained language model (mBERT, XLM-R, or similar) and uses the \texttt{[CLS]} token to classify the utterance. Our Teacher-Student method uses two such models, jointly trained, as outlined in Figure~\ref{fig:model}.
In this setup, the Teacher model acts as an anchor that does not change, i.e. all its multi-head attention layers are frozen. The goal is that the representations produced from training the Student model for the translated and transliterated data will eventually align with the original (Teacher) model's source language pre-trained representations. 

Training consists of two tasks, an unsupervised alignment task and a classification task. The goal of the alignment task is to ensure that the three variants (source, target, and transliterated target) end up with similar representations. As this only requires a 3-way parallel corpus with no labels, it is trained in an unsupervised fashion. The goal of the classification task is to train the model for the given class labels. Joint training on both tasks is necessary for tackling our problem.

Since the two tasks we are interested in only require the classification token for making a prediction, we build our loss functions on top of the \texttt{[CLS]} token. We treat its representation as the encoder's output ($h$), which is directly used to compute the unsupervised alignment loss and passed through linear layers to compute the classification output ($p = \softmax(W h + b)$).
As shown in Figure~\ref{fig:model}, $h_{\sigma}$, $h_{\tr}$, and $h_{\tl}$ denote the representations of $X^{\sigma}$, $X^{\tr(\sigma,\tau)}$, and  $X^{\tl(\tr(\sigma,\tau), \sigma)}$ respectively. 

The unsupervised alignment loss consists of three components: $\mathcal{L}_{ts}$, $\mathcal{L}_{tr}$, and $\mathcal{L}_{tl}$.  The \textit{Teacher-Student Loss} ($\mathcal{L}_{ts}$) is defined as the difference between output embeddings produced using source language ($X^{\sigma}$) by the Student ($s$) versus the Teacher ($t$). 
The \textit{Translation Loss}  ($\mathcal{L}_{tr}$) is defined as the difference between output embeddings produced using $X^{\tr(\sigma,\tau)}$ on the Student versus $X^{\sigma}$ on the Teacher. Similarly, the \textit{Transliteration Loss} ($\mathcal{L}_{tl}$) is defined as the difference between output embeddings produced using $X^{\tl(\tr(\sigma,\tau), \sigma)}$ on the Student versus $X^{\sigma}$ on the Teacher. 

All distances are measured using the cosine similarity between the two vectors.\footnote{We use $d(a,b) = 1 - \frac{a\cdot b}{|a||b|}$.} Essentially, all three losses penalize moving away from the Teacher's representation of the source language: 
\begin{equation}
\small
\vspace{-.5em}
\begin{aligned}
\mathcal{L}_{ts}= \frac{1}{N} \sum &d(h_{\sigma}^{t}, h_{\sigma}^{s}), \hspace{.2cm} \mathcal{L}_{tr} = \frac{1}{N} \sum d(h_{\sigma}^{t}, h_{tr}^{s}) \\
&\mathcal{L}_{tl} = \frac{1}{N} \sum d(h_{\sigma}^{t}, h_{tl}^{s}), \\
\end{aligned}
\end{equation}
\noindent where $N$ represents the number of samples. Thus, the final unsupervised alignment loss ($\mathcal{L}_{u}$) defined over the Teacher-Student model can be defined as the weighted sum of the three losses:
\begin{equation}
\small
\mathcal{L}_{u} = \beta_{1} \mathcal{L}_{ts} + \beta_{2} \mathcal{L}_{tr} + \beta_{3} \mathcal{L}_{tl} \label{eq:4}
\end{equation}
Meanwhile, the loss function ($\mathcal{J}_{joint}$) for the sentiment task is a sum of binary cross-entropy (BCE\footnote{$BCE = - \frac{1}{N} \sum_{i=1}^{N} y_i \log \hat{y_i} +  (1-y_i) \log (1-\hat{y_i})$}) losses, defined similarly for the three variants of parallel data (Source ($\sigma$), Translated ($tr$), and Transliterated ($tl$)): 
\begin{equation}
\small
\mathcal{J}_{joint} = \sum_{k \in [\sigma, tr, tl]} BCE_{k}\label{eq:5}
\vspace{-.3em}
\end{equation}
The overall loss is simply the combination of the unsupervised and the classification loss.
\begin{equation}
\small
\mathcal{L}_{joint\_ts} = \mathcal{J}_{joint} + \alpha \mathcal{L}_{u} \label{eq:6}
\vspace{-.3em}
\end{equation}
where $\alpha$ is the hyperparameter that controls the unsupervised alignment loss. 
To summarize, the Teacher-Student model takes the augmented data ($\mathcal{S'}$) as the input and optimizes over a joint loss function that comprises of an unsupervised component that aligns the translated and transliterated representation into the same vector space as the source language and a supervised component that learns to classify for the task at hand.




\begin{table*} [ht]
\small
\centering
\begin{tabular}{lccccc} 
\toprule
\multirow{2}{*}{\textbf{Dataset}} & \multirow{2}{*}{\textbf{$+$}} & \multirow{2}{*}{\textbf{$-$}} & \textbf{Avg. \# of words}  & \textbf{Avg. \# of chars.}  & \textbf{Avg. \# of chars.} \\
&  & & \textbf{per sentence}  & \textbf{per word}  & \textbf{per sentence} \\
\midrule
\multicolumn{6}{c}{\textbf{Movie Review Datasets - Sentiment}}\\
Hindi Movie Reviews  & 335 & 293 & 154.5 & 4.0 & 613.8  \\
Romanized Hindi ($\hi_{\ro})$ & 335 & 293  & 154.5 & 5.0 & 775.8  \\
Malayalam Movie Reviews & 501 & 451 & 10.4 & 9.3 & 108.2  \\
Romanized Malayalam ($\ml_{\ro})$ & 501 & 451  & 10.4 & 10.8 & 122.6  \\
\midrule
\multicolumn{6}{c}{\textbf{Crisis Tweet Datasets - Relevancy}} \\
North India Floods ($\hi_{nf}$) & 206 & 250 & 20.2 & 3.9  & 78.1   \\
Kerala Floods ($\ml_{kf}$) & 109 & 132 & 19.1  & 5.4 & 103.6   \\
\bottomrule
\end{tabular}
 \caption{Statistics for the test datasets.} \label{table:stat}
 \vspace{-1em}
\end{table*}

\section{Datasets}
In this section, we outline the datasets we use in our experiments.

\subsection{English Movie Reviews}

English movie reviews are sampled from the large IMDb movie review dataset \cite{maas2011learning} with randomly sampled balanced counts of $5000$ positive and $5000$ negative reviews for training and $500$ each for validation. During training, this dataset is translated to Hindi and Malayalam, and subsequently transliterated. Ground truth Hindi and Malayalam datasets (described in the following sections), are used only for evaluation. The `$\ro$' notation used in the following section for representing datasets signify $\tl(\bullet ,\text{en})$; i.e. romanization or Latin-transliteration using transliteration tools.

\subsection{Hindi Movie Reviews ($\hi$, $\hi_{\ro}$)}
\label{hi_data}

The original Hindi movie review dataset\footnote{\url{https://www.kaggle.com/disisbig/hindi-movie-reviews-dataset}} is constructed from various News Websites. This dataset consists of positive, negative, and neutral movie reviews. We select only positive and negative reviews from both training and validation datasets to construct the test dataset for our cross-lingual setup. From this, we construct the romanized Hindi dataset using the Indic-nlp transliteration tool~\cite{kunchukuttan2020indicnlp}.\footnote{\url{https://pypi.org/project/indic-transliteration/}}

\subsection{Malayalam Movie Reviews ($\ml$, $\ml_{\ro}$) - New Dataset I}
\label{ml_data}

For Malayalam movie reviews, we construct a new human-labeled dataset from the Samayam News website.\footnote{\url{https://malayalam.samayam.com/malayalam-cinema/movie-review/articlelist/48225004.cms}} Reviews are lengthy, in general, with plenty of neutral text. So, a native Malayalam speaker was tasked with extracting a few sentences from each movie review such that each positive or negative example in our dataset is highly polar. From this, we construct the romanized Malayalam dataset using the \texttt{ml2en} tool.\footnote{\url{https://pypi.org/project/ml2en/}} A few samples from the dataset are shown in Figure~\ref{fig:ml_stat} and dataset statistics are available in Table~\ref{table:stat}. 

\subsection{Crisis Tweets ($\en$)}
Appen\footnote{\url{https://appen.com/datasets/combined-disaster-response-data/}} provides a labeled collection of tweets posted during various natural disasters such as earthquakes, floods, and hurricanes. The three most common languages in this dataset are English, Spanish, and Haitian Creole. For our experiments, we focus on the \texttt{related} label column signifying the relevancy of tweets. The dataset also contains English translations of non-English tweets. Training and validation datasets are constructed using only the English tweets (\texttt{message} column in the Appen dataset). Statistics are shown in Table \ref{table:stat}.

\subsection{Crisis Tweets: North India ($\hi_{nf}$) and Kerala ($\ml_{kf}$) Floods - New Dataset II}

To construct our test datasets for the crisis tweet classification experiments, we collected tweets from the 2013 North India and 2018 Kerala Floods using Twitter API.  We filtered these tweets to restrict only to naturally-occurring transliterated Hindi and Malayalam sentences, using a set of transliterated crisis-related keywords such as \textit{madad}, \textit{toofan}, \textit{baarish}, \textit{sahayta}, \textit{floods}, etc., for Hindi and \textit{pralayam}, \textit{vellapokkam}, \textit{vellam}, \textit{sahayam}, \textit{durantham}, \textit{veedukal}, etc., for Malayalam. With the help of a native language expert for each language who are also proficient in English, the tweets are labeled based on contextual-relevancy or relatedness; similar to the \texttt{related} label for the English tweets in the Appen dataset. Dataset statistics are shown in Table~\ref{table:stat} and a few tweet samples are shown in  Figure~\ref{fig:ml_stat}.



\section{Experimental Setup}
Since we are interested in cross-lingual transfer, we only use the English datasets (IMDb reviews for Sentiment Analysis and Appen Crisis Dataset for Tweet Classification) for training, augmented with their automatic translations and transliterations.
We evaluate on all other datasets (c.f. Table~\ref{table:stat}).

\paragraph{Monolingual LM Baselines.} 
Our first baseline uses pre-trained language models from Hugging Face \cite{wolf2020transformers} that are monolingually trained on Hindi\footnote{\url{https://huggingface.co/monsoon-nlp/hindi-tpu-electra}} and Malayalam.\footnote{\url{https://huggingface.co/eliasedwin7/MalayalamBERT}}

\paragraph{mBERT/XLM-R Baselines.} 
We also consider baselines using multilingual masked LMs (MLM), specifically mBERT \cite{devlin2018bert} and XLM-R \cite{conneau2019unsupervised}. 
We compare our models with the following MLM baselines: \textbf{1)} a model trained using only in English, \textbf{2)} a model trained using English translated to the target language using MarianMT \cite{mariannmt}, \textbf{3)} a transliterated model which is trained using the target-transliterated 
version of target-translated English data, and \textbf{4)} a combination of the three. These baselines use our augmented datasets but do not use the Teacher-Student training scheme, e.g. $\text{mBERT}_{tl}$ refers to mBERT trained using target-transliterations of target-translated English data.


\begin{table*}[t]
\small
\begin{center}
\begin{tabular}{c*{2}{c}*{2}{c}*{2}{c}*{2}{c}|*{2}{c}}
\toprule
   \textbf{Test Data} $\rightarrow$  & \multicolumn{2}{c}{$\mathbf{hi_{ro}}$} & \multicolumn{2}{c}{$\mathbf{ml_{ro}}$}  & \multicolumn{2}{c}{$\mathbf{hi_{nf}}$} & \multicolumn{2}{c}{$\mathbf{ml_{kf}}$} & \multicolumn{2}{c}{AVG}\\
   
\textbf{Models}  $\downarrow$ & Acc. & F1 & Acc. & F1 & Acc. & F1 & Acc. & F1 & Acc. & F1 \\
\midrule
Monolingual LM$^\diamondsuit$ & 50.38 & 37.99 & 48.76 & 47.55 & 54.39 & 49.06 & 63.35 & 63.33 & 54.22 & 49.48  \\
\multicolumn{9}{c}{\textbf{mBERT Baselines}}  \\
mBERT$_{en}$  & 47.55 & 41.23  & 48.23 & 41.63 & 56.67 & 55.27 & 57.84 & 57.72 & 52.57 & 48.96 \\
mBERT$_{tr}$  & 51.81 & 48.21  & 51.32 & 45.62 & 52.89 & 47.89  & 58.26 & 57.54 & 53.57 & 49.82 \\
mBERT$_{tl}$  & 55.29 & 54.18  & 61.72 & 56.47 & 56.14 & 55.78 & 58.09 & 57.79 & 57.81 & 56.06 \\
mBERT$_{en+tr+tl}$  & 55.16 & 54.75  & 61.72 & 61.25 & 56.71 & 56.03 & \textbf{63.74} & \textbf{63.42} & 59.33 & 58.86 \\
\multicolumn{9}{c}{\textbf{Our mBERT models}}  \\
mBERT-Joint  & 55.57 & 55.56  & 64.29 & 63.58  & 57.75 & 56.73 & 51.85 & 53.98 & 57.37 & 57.46 \\
mBERT-Joint-TS  & \textbf{57.37}$^\spadesuit$ & \textbf{57.36}$^\spadesuit$  & \textbf{65.15}$^\spadesuit$ & \textbf{65.78}$^\spadesuit$ & \textbf{63.22}$^\spadesuit$ & \textbf{63.14}$^\spadesuit$ & 62.55 & 62.40 & \textbf{62.07} & \textbf{62.17} \\
\midrule
\multicolumn{9}{c}{\textbf{XLM-R Baselines}} \\
XLM-R$_{en}$   & 50.57 & 45.76  & 50.86 & 47.13 & 58.11 & 56.77  & 61.74 & 60.98 & 55.32 & 52.66 \\
XLM-R$_{tr}$  & 49.52 & 47.67  & 51.72 & 50.16 & 57.11 & 56.95 & 61.74 & 61.72 & 55.02 & 54.13\\
XLM-R$_{tl}$  & 54.81 & 53.72  & 51.67  & 54.71 & 51.45 & 51.24  & 59.84 & 59.23 & 54.44 & 54.73\\
XLM-R$_{en+tr+tl}$ & 55.57 & 54.19 & 62.46 & 61.52  &  56.40 & 55.67 & 63.24 & 63.10 & 59.42 & 58.62\\
\multicolumn{9}{c}{\textbf{Our XLM-R Models}} \\
XLM-R-Joint  & 56.09 & 55.40 & 62.90 & 63.14 & 53.68 & 52.81  & 62.79 & 62.77 & 58.87 & 58.53\\
XLM-R-Joint-TS & \textbf{57.70}$^\spadesuit$ & \textbf{57.03}$^\spadesuit$ & \textbf{65.93}$^\spadesuit$ & \textbf{65.71}$^\spadesuit$ & \textbf{58.39} & \textbf{57.86} & \textbf{64.87}$^\spadesuit$ & \textbf{64.87}$^\spadesuit$ & \textbf{61.72} & \textbf{61.37}\\
\bottomrule
\end{tabular}
\end{center}
 \caption{Performance evaluation (weighted F1) on various romanized datasets shows that our Teacher-Student model outperforms the baselines.  $\diamondsuit$: Dedicated Hindi and Malayalam language models (LM) from Hugging Face; results reflect the best performing model by varying the training data. $^\spadesuit$: The difference is significant with p $<$ 0.05 using Tukey HSD (compared against the best baseline model).} \label{table:scores}
\end{table*}


\begin{table}[t]
\small
\centering
\begin{tabular}{cc|cc}
\toprule
   \textbf{Train Language} $\rightarrow$ & $\mathbf{en}$ & $\mathbf{hi}$ & $\mathbf{ml}^\dagger$      \\
\textbf{Model} & (Baseline) & \multicolumn{2}{c}{(Joint-TS Model)}\\
\midrule
mBERT & 81.06 & 82.35  & 71.25 \\
XLM-R & 83.47 & 84.00  & 74.16    \\ 
\bottomrule
\end{tabular}
 \caption{Evaluation (weighted F1) on IMDb English test data showing that our model preserves the performance on the source language for Hindi but not for Malayalam. The results in $\hi$ and $\ml$ are with our modifications. $\dagger$: See discussion.} \label{table:scores2}
\end{table}

\begin{table}[t]
\small
\begin{center}
\begin{tabular}{ccc}
\toprule
   \textbf{Test Language} $\rightarrow$   & $\mathbf{hi}$ & $\mathbf{ml}$     \\
\midrule
mBERT$_{\en}$ & 54.34  &  54.07  \\
mBERT$_{\tr}$ & 60.11 &  66.20  \\
mBERT-Joint-TS  & 61.35  & 66.24 \\
\midrule
XLM-R$_{\en}$ &  60.82 & 78.11 \\
XLM-R$_{\tr}$ &  59.72 & 71.40 \\
XLM-R-Joint-TS & 62.23 & 76.46$^\dagger$\\
\bottomrule
\end{tabular}
\end{center}
 \caption{Evaluation (weighted F1) on Hindi and Malayalam original script showing that our model preserves or outperforms their performance on the baselines except for Malayalam on XLM-R. $\dagger$: See discussion.} \label{table:scores3}
\end{table}

\paragraph{Our Proposed Models.} 

\textbf{Joint-TS} represents our Teacher-Student model shown in Figure \ref{fig:model} and Eq.~\ref{eq:6}. We also perform an ablation (the \textbf{Joint} model) without the Teacher model (setting $\alpha=0$ in Eq.~\ref{eq:6}), i.e. this model does not have an anchored Teacher, which means that there is no penalty for representations of parallel sentences being different. 

\paragraph{Implementation Details.}

Our implementation is in PyTorch  \cite{NEURIPS2019_9015} with the transformers library~\cite{wolf2020transformers}. We use the pre-trained cased multilingual BERT and the pre-trained \textit{xlm-roberta-base} with a standard sequence classification architecture. 
Maximum epoch is set to~$40$ with an early stopping patience of~$10$, batch size to~$32$, and we use the Adam optimizer \cite{kingma2014adam} with a learning rate of~$5\mathrm{e}{-5}$. We select the best model based on the validation set, using both accuracy and weighted F1 as performance measures.

\section{Results \& Discussion}
\label{results}


\paragraph{Performance on Transliterated Target.} Table \ref{table:scores} shows the classification performance\footnote{Model runtime shown in Appendix \ref{app:B}.} on transliterated datasets. Averaging the scores, we see a total boost of $+5.6\%$ on mBERT and $+4.7\%$ on XLM-R in weighted F1 performance across the 4 romanized datasets.\footnote{With similar trends on accuracy as well.}  
Our top baseline models in each of the masked language models are mBERT$_{en+tr+tl}$ and XLM-R$_{en+tr+tl}$, which combine the augmented data with the original data to fine-tune the model for the classification task. The improvement produced by our model is due to the fact that the pre-trained models have likely not seen that many transliterations in these languages. mBERT is trained using data from Wikipedia, while XLM-R uses Common Crawl. As such, XLM-R likely has been trained on at least some code-switched or transliterated data. This is also supported by the scores in Tables~\ref{table:scores}, \ref{table:scores2}, and \ref{table:scores3}, where XLM-R$_{en}$ outperforms mBERT$_{en}$ on all datasets, i.e. XLM-R's English representations are much more generalizable than mBERT's for these two languages.

\begin{figure}[t]
    \centering
    \includegraphics[scale=0.18]{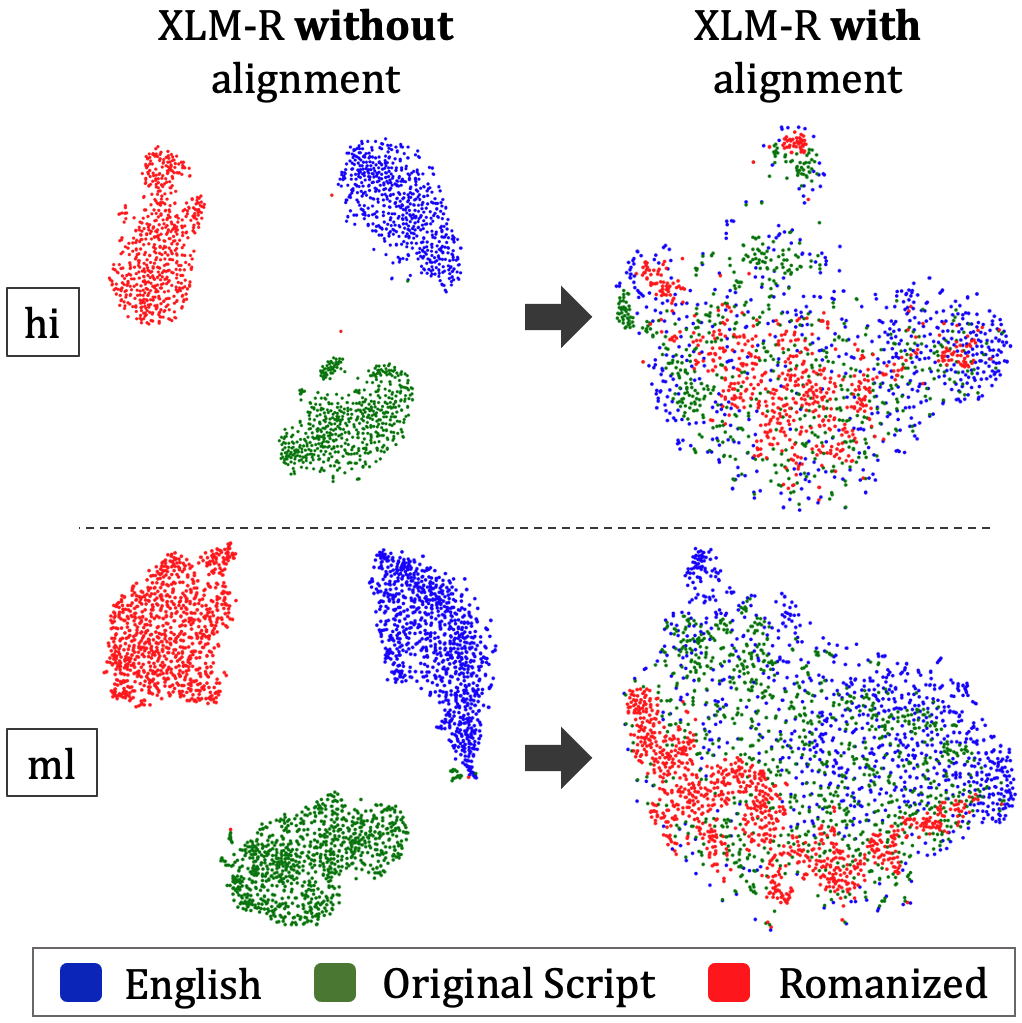}
    \caption{The visualization of final layer representations (from the movie review test data) shows the effectiveness of unsupervised alignment training in bringing the representations of the 3 variants in a shared space. } \label{fig:unsup}
    \vspace{-1em}
\end{figure}

\paragraph{Performance on Original-Script Target.} Beyond improving on transliterated text, we need our model to handle original, not-transliterated text equally well. Evaluating on the original script (Table~\ref{table:scores3}) we find that our models do preserve or even outperform the baselines trained only with English or non-romanized text, especially in Hindi. 
The exception is Malayalam in XLM-R, which could be attributed to the power of XLM-R pre-training in producing a more generalizable English representation than the original script ($\text{XLM-R}_{\en}$ vs. $\text{XLM-R}_{tr}$).

\paragraph{Performance on English.}  Ideally, our Teacher-Student model will preserve the performance on the source language  (in this case English).  Table~\ref{table:scores2} shows the performance evaluation on English data ($1000$ randomly selected positive and negative reviews from the IMDb dataset \cite{maas2011learning}). We observe that fine-tuning on Hindi preserves and slightly improves performance in English, but Malayalam does not ($\dagger$ in Table~\ref{table:scores2}). We speculate that, for Malayalam, this is because the transliterated embeddings (red) are not blended in well with the others compared to those for Hindi as shown in Figure~\ref{fig:unsup} (right panels).

\paragraph{Representation Alignment.} Figure~\ref{fig:unsup} shows a visual reason behind our model's aforementioned performance improvements. While the t-SNE projection of the (3-way parallel) sentence representations for the original model are quite distinct based on language/script (left), our model brings all representations in the same vector space (right). This method of unsupervised training ($L_{u}$ in Eq. \ref{eq:4}) is very generic in nature that it can be adapted for any alignment scenarios where different variants or dialects of the same language need to be aligned.

\paragraph{Ablation Studies.} We conduct $8$ ablations in total. First four mBERT/XLM-R models in Table~\ref{table:scores} are ablations that show how the three language variants and their combination perform. The joint model without Teacher-student in Table~\ref{table:scores} represents the version without the alignment training. Table~\ref{table:scores2} shows performance on English and Table~\ref{table:scores3} shows performance on original script. The impact of unsupervised alignment is shown in Figure~\ref{fig:unsup} showing where the boost is coming from. Ablation of the Joint-TS model without $tr$ is not shown as our goal is to construct a single model that preserves performance across the three variants.




\begin{figure} [t]
    \centering
    \includegraphics[scale=0.42]{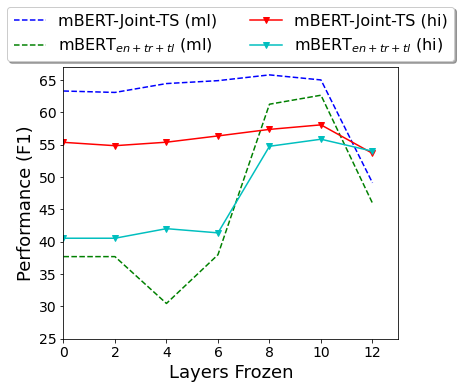}
    \caption{Freezing up to layers $8$ to $10$ of mBERT showed to be optimal for our task.} \label{fig:freeze}
    \vspace{-1em}
\end{figure}

\paragraph{Unsupervised Domain Adaptation.} In addition to being cross-lingual, our approach also falls under the paradigm of unsupervised domain adaptation. Our sources and targets do not strictly fall in the same domain, even though the classification tasks are similar. For example, the training data for classifying tweets consists of not only floods, but also other events such as hurricanes, fires, earthquakes, etc. This leads to another application of our method, which can be deployed at the onset of crisis events in any region, with minimal requirement to collect labeled data from the new crisis, or  
converting the data to native script or English,  
which might save precious time for crisis response.

\paragraph{Hyperparameter Tuning.} Primarily, we tune two hyperparameters: a) $\alpha$ - the weight that is given to unsupervised alignment loss in Eq. \ref{eq:6} and b) number of layers to freeze to identify the appropriate amount of pre-trained information to be preserved without alteration. $\alpha$ values are tuned using a simple grid search from a range of [$0.01$-$1.0$]. All $\beta$ values (Eq. \ref{eq:4}) are set to $1$ to prioritize the three variants (source, target, transliterated target) equally. For Joint-TS (Eq. \ref{eq:6}), best hyperparameters for mBERT are $\alpha_{\hi_{\ro}}$= $0.3$ and best $\alpha_{\ml_{\ro}}$= $0.05$ and for XLM-R are $\alpha_{\hi_{\ro}}$= $0.5$ and best $\alpha_{\ml_{\ro}}$= $0.01$. Intuitively, we find that giving the classification loss primary focus and the unsupervised alignment loss secondary focus produced better results. On the other hand, we found that freezing bottom layers 
leads to better results, as shown in Figure~\ref{fig:freeze}. The optimal amount of layers to freeze was empirically between $8$ to $10$. 


\section{Related Work}

Sentiment analysis in Hindi spans a variety of tasks such as analysis of movie reviews \cite{nanda2018sentiment}, building subjective lexica for product reviews and blogs \cite{arora2013sentiment}, analysis on tweets \cite{sharma2015practical}, aspect-based sentiment \cite{akhtar2016aspect}, predicting elections~\cite{sharma2016prediction}, and analysis on code-mixed text~\cite{joshi2016towards}. Code-mixing is another related phenomenon where multilingual speakers alternate between languages, often in the same sentence. Both code-mixing and transliteration are studied for Hindi and Marathi texts using supervised learning methods by \citet{ansari2018sentiment}. We restrict our analysis to transliteration, although our dataset may contain code-mixed text. Recently, \citeauthor{khudabukhsh2020harnessing} have proposed a pipeline to sample code-mixed documents using minimal supervision. In cross-lingual context, researchers have used linked WordNets~\cite{balamurali2012cross} and cross-lingual word embeddings~\cite{singh2020sentiment} using MUSE~\cite{conneau2017word} and VecMap~\cite{artetxe2018robust} to bridge the language gap, later addressing code-mixing and transliteration. With the advent of large pre-trained language models, we take a step further in this direction to enhance mBERT/XLM-R to cover transliterations for fine-tuning it to the downstream task of sentiment analysis.

Malayalam has also seen several works on sentiment analysis~\cite{nair2015sentiment,kumar2017sentiment,nair2014sentima,ashna2017lexicon}. Recently, a new Malayalam-English code-mixed corpus~\cite{chakravarthi2020sentiment} has been constructed by scraping YouTube comments. This corpus primarily consists of romanized sentences with some code-mixing. After converting this dataset into its original script to obtain the parallel corpus, this can also be used as an additional dataset for our model evaluation.

Translate and train has been a popular methodology \cite{shah2010synergy, yarowsky2001inducing, ni2017weakly, xu2020end} that utilizes the power of existing Machine Translation tools \cite{wu2016google,mariannmt} to perform cross-lingual tasks by augmenting the original source dataset with the target-translated data before training. This kind of training could enhance the performance of multilingual representations by fine-tuning the pre-trained models such as mBERT or XLM-R, creating a pseudo-supervised environment where the model now has access to data in the target language. We follow the same approach to create strong baselines as well as for the Teacher-Student model.

\section{Future Work}
We acknowledge that existing Hindi and Malayalam translation and transliteration tools pose limitations and may cause cascading errors. For example, the Indic transliteration tool adds an extra, unnecessary \textit{`a'} for some Hindi words (eg., \textit{`sandarbh'} vs \textit{`sandarbha'}) which may not reflect how native users tend to write. Despite its limitations, using it is still beneficial to construct a strong baseline. We expect that improved transliteration systems would further improve downstream accuracy. This is also the case for the many Indic languages where extensive datasets are not available. We plan to expand our model to other Indic languages such that their translations and transliterations are aligned within the language models such as mBERT/XLM-R. Another future direction is to perform an in-depth sensitivity analysis over the $\beta_{2}$ and $\beta_{3}$ parameters that tune the unsupervised alignment loss ($L_{u}$) in Eq.~\ref{eq:4}, which might also address some of the exceptions shown in Tables~\ref{table:scores2} and \ref{table:scores3}, to ensure that the model preserves performance on English or non-romanized target data.

\section{Conclusion}
We propose a Teacher-Student model to enhance the multilinguality of language models such as mBERT/XLM-R so that it can be adapted to perform cross-lingual text classification tasks for \textit{transliterated} Hindi and Malayalam. Experiments show that our model outperforms traditional fine-tuning and other baselines built on the state-of-the-art. Furthermore, we release two human-annotated datasets: a highly polar Malayalam movie review dataset for sentiment analysis and a dataset of Hindi and Malayalam romanized tweets posted during North India and Kerala floods. 
Additionally, our method presents a generic and extensible architecture that could be adapted to any language alignment scenarios where large pre-trained multilingual models may fall short.\\ 


\section{Acknowledgement}
We thank U.S. National Science Foundation grants IIS-1815459 and IIS-1657379 for partially supporting this research. We thank Ming Sun, Alexis Conneau, Sneha Mehta, and Raj Patel for giving valuable insights on language models, machine translation, and multilingual model training. We thank Chandini Narayan and Sujay Das for data annotation. We also acknowledge ARGO team as the experiments were run on ARGO, a research computing cluster provided by the Office of Research Computing at George Mason University.

\bibliography{anthology,acl2021,custom}
\bibliographystyle{acl_natbib}
\clearpage
\newpage
\newpage

\appendix

\section{Alignment Deficiency}
\label{app:A}
\begin{figure*}[hbt!]
    \centering
    \includegraphics[scale=0.25]{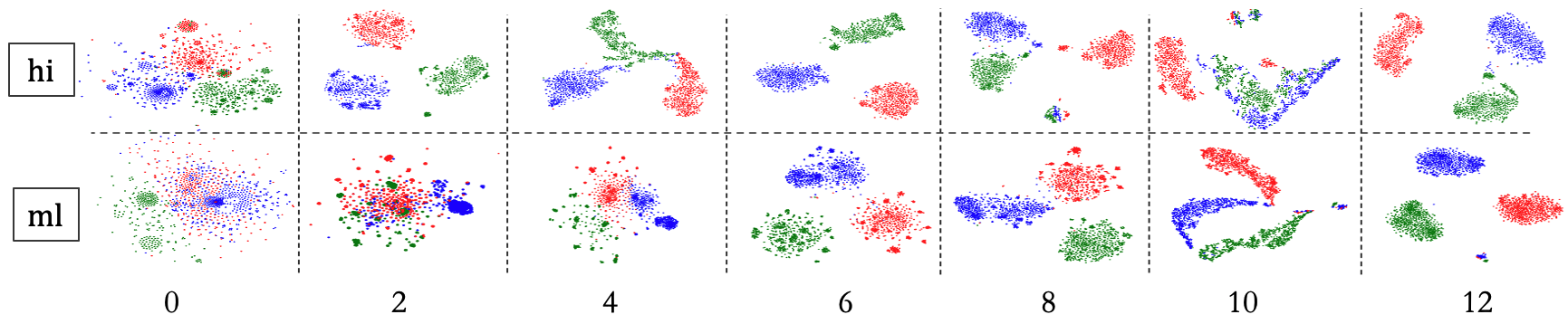}
    \caption{t-SNE plot of embeddings across the multi-head attention layers shows that the alignment deficiency identified in previous works \cite{pires2019multilingual} also extends to transliteration. XLM-R on Hindi (top). mBERT on Malayalam (bottom). English (blue), Original Script (green), and Latin-Transliteration (red).} 
    \label{fig:tsne}
\end{figure*}

Continuing the discussion from the introduction (Section \ref{intro}), Figure~\ref{fig:tsne} describes the representational deficiency exhibited by mBERT and XLM-R on Hindi and Malayalam.  The t-SNE plot of 3-way parallel datasets consisting of sentences in source (English), target (Malayalam/Hindi), and their romanizations are clearly clustered in their own groups, even though they match semantically. Our goal is to address this issue and bring them into a comparable vector space as shown in Figure~\ref{fig:unsup} using a Teacher-Student training scheme.

\section{Model Runtime}
\label{app:B}
As an addendum to the performance evaluation shown in Table~\ref{table:scores}, we also provide the runtime of our Joint-TS model and key baselines. The $\text{mBERT}_{\en}$ and $\text{XLM-R}_{\en}$ model are trained only on English data without any augmentation while $\text{mBERT}_{\en+tr+tl}$, $\text{XLM-R}_{\en+tr+tl}$, and the Joint-TS models are trained using translated and transliterated target in addition to English data. The additional latency is caused due to the augmented data (two times more data). Our Joint-TS model also consists of unsupervised optimization for alignment, in addition to the augmented data. An interesting observation is that the Teacher-Student model based on mBERT converges faster than its ${\en+tr+tl}$ counterpart and XLM-R, while also having a comparable performance as shown in Table~\ref{table:scores}.

\begin{table*}
\small
\begin{center}
\begin{tabular}{ccccccc}
\toprule
   \textbf{Model} $\rightarrow$   & mBERT$_{en}$ & XLM-R$_{en}$ & mBERT$_{en+tr+tl}$  & XLM-R$_{en+tr+tl}$ & mBERT-Joint-TS & XLM-R-Joint-TS  \\
\midrule
\textbf{Runtime} $\rightarrow$  & 00:35:55 & 00:55:53 & 02:38:39 & 02:40:48 & 01:30:48 & 04:39:16 \\
\bottomrule
\end{tabular}
\end{center}
 \caption{Run time on a single K80 GPU in HH:MM:SS for training a Malayalam model on Crisis data.} \label{table:runtime}
\end{table*}

\section{Ethical Considerations}
\label{app:ethics}
The tweets extraction procedure followed the Twitter Terms of Service and did not violate privacy policies of individual users. Also, the datasets we share 
include only Tweet IDs in the public domain. Data statement that includes annotator guidelines for the labeling jobs and other dataset information will be provided with the implementation. From a broader impact perspective, our code is open-source and allows NLP technology to be accessible to information systems for emergency services and social scientists in studying a large population in India who use transliterated text for communication in everyday life.

\end{document}